\documentclass[letterpaper]{article} 
\usepackage{aaai24}  
\usepackage{times}  
\usepackage{helvet}  
\usepackage{courier}  
\usepackage[hyphens]{url}  
\usepackage{graphicx} 
\usepackage{natbib}  
\usepackage{caption} 
\usepackage{algorithm}
\usepackage{algorithmic}

\usepackage{newfloat}
\usepackage{listings}
\DeclareCaptionStyle{ruled}{labelfont=normalfont,labelsep=colon,strut=off} 
\floatstyle{ruled}
\newfloat{listing}{tb}{lst}{}
\floatname{listing}{Listing}
\title{What Makes An Expert? Reviewing How ML Researchers Define ``Expert''}
\author{
    Mark D\'iaz\textsuperscript{\rm 1},
    Angela D.R. Smith\textsuperscript{\rm 2}
}
\affiliations{
    \textsuperscript{\rm 1}Google Research\\
    \textsuperscript{\rm 2}University of Texas at Austin\\ 

    markdiaz@google.com, adrsmith@utexas.edu
}

\begin{document}

\maketitle

\begin{abstract}
Human experts are often engaged in the development of machine learning systems to collect and validate data, consult on algorithm development, and evaluate system performance. At the same time, who counts as an ‘expert’ and what constitutes ‘expertise’ is not always explicitly defined. In this work, we review 112 academic publications that explicitly reference ‘expert’ and ‘expertise’ and that describe the development of machine learning (ML) systems to survey how expertise is characterized and the role experts play. We find that expertise is often undefined and forms of knowledge outside of formal education and professional certification are rarely sought, which has implications for the kinds of knowledge that are recognized and legitimized in ML development. Moreover, we find that expert knowledge tends to be utilized in ways focused on mining textbook knowledge, such as through data annotation. We discuss the ways experts are engaged in ML development in relation to deskilling, the social construction of expertise, and implications for responsible AI development. We point to a need for reflection and specificity in justifications of domain expert engagement, both as a matter of documentation and reproducibility, as well as a matter of broadening the range of recognized expertise.
\end{abstract}

\section{Introduction}
Human expertise plays a critical role in various facets of machine learning (ML) development. Human experts are often engaged in the development of ML systems to collect and validate data, consult on the development of knowledge representations, and evaluate system performance. For systems designed for use in highly specific contexts, such as predicting car insurance fraud \cite{vsubelj2011expert}, domain experts provide knowledge that others involved in development may lack. Indeed, experts can play an important role as a reference for guiding the behavior of a model, for example by providing input as ground truth to be be used in model testing. In addition, experts provide context about a given problem domain to identify relevant variables and measures and shape algorithm design. 

At the same time, who counts as an ‘expert’ and what constitutes ‘expertise’ is not always clear or explicitly defined and can shift in relation to system goals. For example, experts in different fields may disagree on optimal solutions or on different approaches toward achieving the same solution. Input from domain experts plays a significant role in ML development as a basis for systems designed to match or exceed human ability in a given set of tasks. However, there exist many kinds of knowledge and expertise, raising questions about the range and consistency of definitions of expertise in ML research. For example, in data annotation, ``expert'' has been used to refer to domain knowledge rooted in annotator lived experience \cite{patton2019annotating}, domain knowledge based on specific training, such as Wikipedia article editing or academic study \cite{kittur2008crowdsourcing, sen2015turkers}, and even to refer to labels from gold standard datasets, even when the gold standard annotators themselves are unspecified \cite{snow2008cheap}.

Understanding how expertise is defined is important, because the label “expert” directs attention to whose knowledge should be upheld as canonical; thus, \textit{who} or \textit{what} is identified as ‘expert’ carries implicit, normative claims about whose knowledge and experience is valid and trustworthy for system development. Thus, expertise confers power to those identified as possessing it \cite{handy1976so}.

In addition to questions of who is included and excluded in ML development, are issues related to \textit{how} domain experts are engaged in ML development. \citet{birhane2022power} point out that participation in ML development can bring with it varying degrees of extractive power relations, particularly when central goals are performance-driven. While Birhane et al. primarily focus on community-based and participatory engagements, participation on the part of a range of subject matter experts is also subject to these dynamics. Other scholars have discussed ways in which expertise has been devalued through automation and ML work despite being sought and acknowledged as specialized and critical \cite{sambasivan2022deskilling, duran2021deskilling}. The potential for extractive and devaluing engagements underscores a need to consider not just how expertise is defined, but also the social and power relations between experts and the development processes that engage them.

Based on a systematic review of ML publications, we contribute a taxonomy detailing the ways ML research studies treat the terms ``expert'' and ``non-expert'' and how recognized experts are engaged in ML development. We find that ML research often leaves expertise undefined and rarely seeks forms of knowledge outside of formal education and professional certification, which has implications for the kinds of knowledge that are recognized and legitimized in and through ML development. Moreover, we find that expert knowledge tends to be utilized in ways focused on mining textbook knowledge-- or discrete information that can be memorized and reproduced relatively easily, such as through data annotation. On the backdrop of calls for expanded participation in ML development, we discuss the social embeddedness of expertise, patterns in ML research in relation to deskilling and power, and the need for more equitable recognition of expertise in support of responsible ML practices.

\section{Related Work}

In this work, we investigate domain experts' role in ML through a sociotechnical lens. As such, we consider the interactions between domain experts and the ML systems and processes they participate in, as well as the knowledge they possess that development teams seek to capture. Notably, there are contrasts in the study of expertise between ML and the social sciences that raise questions about the role of domain expert knowledge in ML development. 

\subsection{Characterizing Knowledge}

A significant amount of work in ML and AI is motivated by a desire to emulate and extend human abilities. For example, expert system development, an early-established sub-field within AI, is dedicated to developing systems that emulate human decision making and reasoning. Critical components to the development of expert systems are knowledge bases and inference engines, which encapsulate the facts and logical rules relevant to a given problem space \cite{bobrow1986expert}. In seeking to enhance knowledge bases, researchers in ML have engaged domain experts to various degrees to help develop these system components-- often through consultations in order to procure inputs to knowledge bases or to validate system information. Expert systems are just one focus within AI, but provide an example of how domain experts have contributed to AI development.

The process of working with domain experts to develop AI systems has been acknowledged as a challenge for several reasons, including the learning curve that data scientists and ML experts face in familiarizing themselves with a new domain, the time and financial costs associated with gathering input from domain experts, and the gap between knowledge as articulated by a domain expert and information as interpretable by a ML system \cite{park2021facilitating}. As a result, much work has focused on developing better processes for externalizing the knowledge of experts for distillation into AI systems. For example, \citet{pomponio2014reducing} developed a methodology to address challenges inherent in supporting domain experts to externalize tacit knowledge, and other researchers have similarly developed approaches to more easily elicit and make us of expert knowledge \cite{rizzo2018harnessing, hobballah2018formulating, steenwinckel2021flags, zhitomirsky2017toward}.

\subsubsection{A Social Science View of Knowledge} Underlying much work focused on eliciting knowledge from domain experts is a conceptualization of knowledge as a discrete entity to be extracted from expert sources and integrated into automated systems \cite{forsythe1993engineering}. In contrast, work across the social sciences--and HCI in particular-- has both questioned the nature of expertise and the degree to which knowledge can be extracted from individuals at all. These conceptualizations characterize knowledge as complex, hard to pin down, and in flux. In ethnographic work observing and interviewing expert system developers, Diana Forsythe comments on a view implicit in how expert systems are developed, wherein knowledge is understood to be rule-based and universally applicable across contexts. She notes this is reflected in the common practice of consulting a single or small set of experts. On the other hand, social scientists view knowledge as socially and culturally constituted \cite{forrester1961industrial}.

Like work in ML, some researchers in HCI have pursued design processes to facilitate knowledge extraction for users or for integration into systems \cite{park2021facilitating, abranches2019nurse}. We draw on social scientists whose work has informed and been expanded upon in HCI research to further articulate knowledge and its characterization. In particular, HCI has investigated modeling human expert knowledge in a variety of ways, such as through mining techniques to determine the breadth and depth of user knowledge in a StackExchange community \cite{kumar2016mining}. In other work, researchers identified topic experts on Twitter in order to surface trustworthy news stories \cite{zafar2016wisdom}. A range of work has explored collective knowledge and crowd innovation (e.g., \cite{yu2016encouraging}), such as through research focused on managing crowdwork feedback for design \cite{chan2016improving, xu2014voyant, luther2015structuring}. Importantly, researchers have also critiqued missing specificity in the use of terms describing expertise. Burns et al. analyzed publications on visualization, noting inconsistencies and disagreements across papers with respect to how "novice" is defined \cite{10.1145/3544548.3581524}. In generating a better understanding of ML development processes and participation, we pursue a similar analysis.

Critically, a number of social scientists have turned focus to social dimensions of expertise, including where expertise "lives" and how it is recognized by others. For example, social dynamics, such as one's distance from others within a social network, can shape the kind of expertise people seek and how they engage with an expert \cite{ehrlich2007searching}. Powell et al. argue that expertise does not exist among isolated experts but rather exists in social and organizational context, such as in the case of tacit knowledge \cite{powell1993experts}. This stands at odds with methods of consulting experts as repositories of knowledge in system development. In this vein, \citet{sambasivan2022deskilling} explored how AI developers recognize and characterize worker expertise. They found that developers failed to recognize domain expertise, treated workers as non-essential, and framed them simply as data collectors despite their unique insights into data collection and problem formulation challenges. Expanding from Sambasivan et al.'s qualitative work, we contribute a review of ML publications to understand the relationship between developers and experts across a variety of ML contexts.

\subsection{Values Embedded in Knowledge}
Domain experts are one of a range of stakeholders that might be involved in the development of an ML system to shape its performance. A growing body of work in AI calls for expanded recognition of the knowledge that different stakeholders possess, as well as broadened participation of oft-ignored stakeholders in the development of systems (e.g., \cite{birhane2022power, bondi2021envisioning, martin2020participatory}). Reasons for this include distributing power over systems that stand to impact communities disparately \cite{birhane2022power}. In particular, scholars have identified the participation of different stakeholders as a means to align system values with a range of stakeholder values \cite{liao2019enabling}. A significant body of work has explored the values that shape AI systems, including the values that motivate and set the goals of AI research \cite{birhane2021values} and differences between the values expressed by AI developers and those of the broader public who stand to be impacted by AI system decision making \cite{jakesch2022different}.

Like other stakeholders, domain experts bring value-laden perspectives shaped by their experiences, social contexts, and communities of practice. For example, \citet{birhane2021values} investigated values expressed in ML research publications, finding a pattern of motivations, such as generalization, novelty, and efficiency. In addition, researchers have found variations in expert judgments from different communities, showing significant differences in the annotations they provided for a hate speech dataset \cite{sen2015turkers}. We adopt a social science view of expertise and affirm that experts relied upon for objective input actually apply value-laden perspectives. Moreover, different expert communities can reflect values that differ or overlap to varying degrees. As Ribes and Bowker show, the development of agreed upon knowledge within a domain is iterative, social, and reflective of the norms that shape what is considered to be legitimate knowledge within that domain \cite{ribes2009between}. This work raises questions about how to negotiate differences in expert perspectives, as well as how the reliance on chosen experts should be justified when expertise from more than one domain may be relevant.

Beyond professionals certified through training or education in a given domain, there exists a landscape of values among individuals and communities that might be consulted for expert input. Indeed, researchers advocating for and pursuing participatory AI are explicitly interested in shaping system development in this way, such as in \citet{zhu2018value}'s creation of a recommender system through the participation of the Wikipedia editors it was intended to serve. From AI fairness and responsible AI (RAI) perspectives, these values are important to understand because the knowledge and expertise captured in data-driven systems can shape how they operate. As researchers increasingly acknowledge the lack of singular ground truth \cite{aroyo2013crowd, davani2021dealing, prabhakaran2021releasing, diaz2022crowdworksheets}, important questions emerge about \textit{which} judgments should be captured in datasets and systems. Intuitively, the individuals and communities that are recognized as expert in a given domain are references for validated information. However, \citet{friedman1996bias} underscore the role of pre-existing societal biases that complicate this intuition. Societal biases shape what society deems to be legitimate forms of knowledge-- including the values and perspectives that these legitimized forms of knowledge reflect. Moreover these values and knowledge can become embedded in ML data and processes without contestability as to how they impact model reliability \cite{smart2020why}.

Ultimately, even traditional notions of expertise considered to be objective reflect disciplinary and domain values, and the range of non-traditional experts (as highlighted in participatory design) bring further perspectives and values that are important to include in RAI practices. The systematic review that we contribute gives us insight into the degree to which ML researchers currently recognize the range of values across domains and the stakeholders within them.


\subsubsection{Reviews of Experts in ML}

Importantly, a few prior surveys have reviewed the roles of domain knowledge in ML development. \citet{von2021informed}, for example, taxonomized knowledge integration into ML systems, investigating which knowledge sources were used by developers and in which stages of the learning pipeline they were integrated. They characterized knowledge sources and how knowledge is formally represented within ML models. Our approach complements that of von Rueden et al. by more narrowly focusing on definitions of expertise articulated in ML research and simultaneously broadening scope to include engagements with knowledge that is not formally represented in a model. This includes engagements that do not directly shape knowledge representation, such as in testing and evaluation, as well as engagements in which experts are consulted for background knowledge on a problem domain.


In addition, other work has investigated active learning and human-in-the-loop (HITL) approaches, which rely on human input in training and testing models. Wu et al. \cite{wu2022survey} surveyed a range of papers describing HITL approaches to ML. The work summarizes the state of HITL work in ML, noting how HITL approaches have been used to address various tasks in natural language processing and computer vision. The summary includes an overview of the different roles humans have played in development, such as "users" or "collaborators", though human engagement in HITL approaches is generally limited to interaction with systems already developed and, while the authors describe the role of experts in general HITL workflows, they do not break down their review by determinations of human expertise. A related review of HITL approaches by Budd. et al. includes explicit discussion of experts but these discussions are limited to annotation processes and applications for medical image analysis \cite{budd2021survey}.

We build on prior work by considering expert participation throughout the entire ML development pipeline, from problem formulation and system evaluation. We call into question not only the roles that experts play in development, but also how human expertise is conceptualized. Critically, existing reviews of domain expert involvement in ML development largely focus on a view of knowledge formalized through quantitative means and applied universally. They do not analyze the assumptions that underlie what constitutes expertise, nor do they discuss the social and ethical significance of how expertise knowledge is defined -- particularly in the frame of a social view of knowledge.

\begin{table}
\centering
\begin{tabular}{cc}
\multicolumn{2}{c}{\textbf{Top Bigrams}} \\ \hline
gold standard & domain expert \\ \hline
expert system & expert systems \\ \hline
domain expertise & domain expert \\ \hline
non expert & expert knowledge \\ \hline
domain specific & machine learning \\ \hline
-based expert & fuzzy expert \\ \hline
gold standards & case study \\ \hline
non experts & human expert \\ \hline
\end{tabular}
\caption{Initial list of top bigrams}
\end{table}

\begin{table}
\centering
\begin{tabular}{cc}
\textbf{Final Keywords} & \textbf{Count} \\ \hline
domain expert* & 175 \\ \hline 
expert knowledge & 69 \\ \hline 
non expert* & 66 \\ \hline 
human expert* & 31 \\ \hline 
\end{tabular}
\caption{*includes all word endings}
\label{keywords}
\end{table}


\begin{figure*}[!ht]
\centering
\includegraphics[scale=.42]{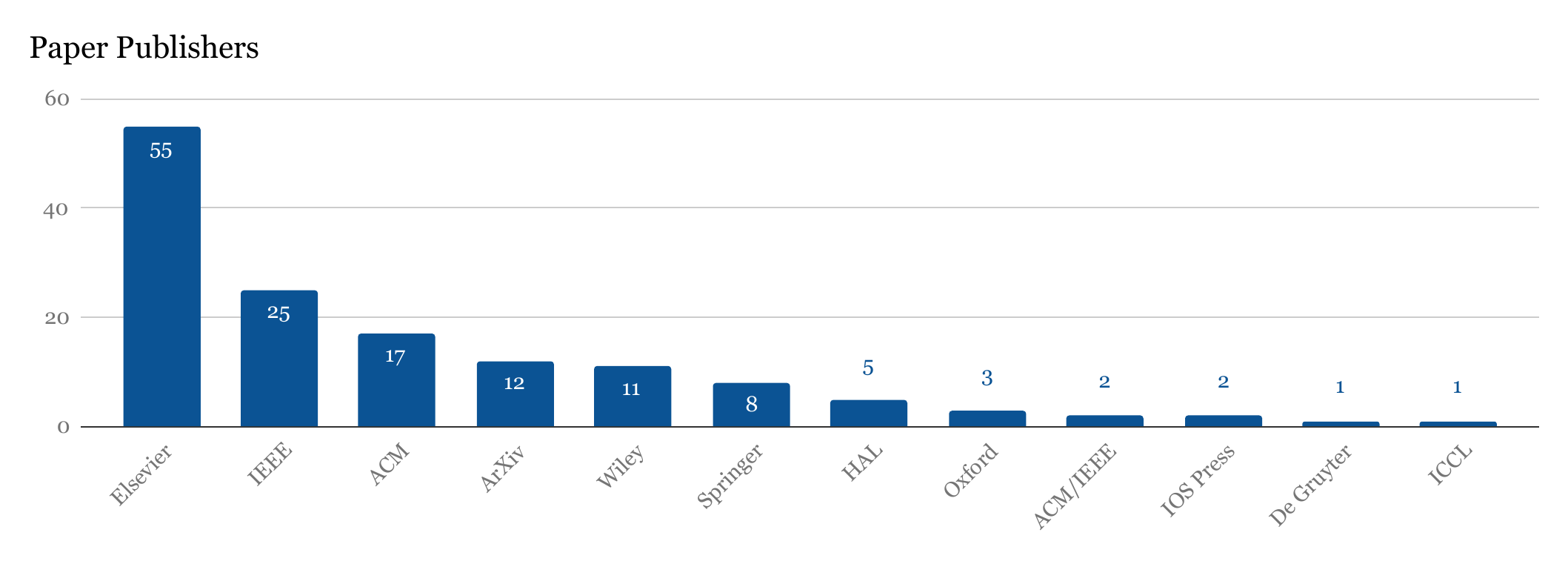}
\caption{Paper count in the final set, by publisher. ICCL refers to the International Committee of Computational Linguistics.}
\label{publishers}
\end{figure*}

\section{Method}
To characterize definitions of expertise and how it is utilized in ML development, we conducted a systematic literature review and thematic analysis \cite{clarke2015thematic, braun2006using} of ML publications that involve domain experts in the development of a ML system. 

\subsection{Systematic Search}
We chose dblp.org to conduct our search. dblp.org is a bibliography of over 6 million computer science publications from a variety of venues, including prominent ML conferences such as NeurIPS, ICML, and AAAI, as well as venues with broad interdisciplinary representation such as CHI and FAccT. The breadth of the bibliography allowed us to capture a wide range of work in ML while also including work that is intersectional in nature.

Before the primary search, we collected results in a general search using the terms “expert”, ”expertise”, and “domain expert” to calculate the top occurring bigrams, which we then used as key words for the full search. Deriving key words in this way allowed us avoid constraining search results as a result of potential author ignorance regarding relevant phrases or terms of art used in ML sub-fields. We selected the top 15 bigrams, though ultimately removed the majority during the course of coding due to the prevalence of irrelevant results they produced. For example, the top initial bigram, ``gold standard'' produced papers that very rarely made mention of human expertise.

The final list of bigrams we generated (after pruning unhelpful keywords) is shown in Table \ref{keywords} along with the number of non-unique paper titles in which the terms appeared. In our systematic search, we focused on references to experts and expertise. Though not the primary focus of our analysis, we kept keywords related to non-experts in order to enable complementary analysis of how non-experts are engaged in ML research. The full keyword list left us with 712 publications before we iteratively removed keywords.

A limitation of using dblp.org is that the service only searches paper titles. We alternatively considered conducting searches of individual conference proceedings and journals and additionally considered search services with broader reach, such as Google Scholar. However, we found that dblp.org included a wider set and variety of publication venues than we initially compiled, in addition to fewer unrelated and off-topic search results. Ultimately, we elected to pursue results that were more relevant across a variety of domains rather than complete comprehensiveness.

We next conducted a full pass, randomly splitting publications between the authors and open coding publications. 10\% of the publications were coded by both researchers to assess the relevance of the keywords we derived and to discuss initial codes before continuing with the full coding pass. For each paper, we open coded 1) the definition of expertise used if one was explicitly provided (if none was provided, we noted an implicit definition), 2) the phase(s) of development experts or non-experts contributed to, as well as 3) their role or how they participated. It was during the initial and full coding passes that we removed keywords from our list, including “gold standard(s)”, “fuzzy expert”, and “expert system(s)”, because these terms produced a large volume of publications that rarely referenced human experts or expertise, and “expert system(s)” specifically drew papers from across the sub-field of expert systems, the vast majority of which did not meet inclusion criteria. This subset of papers describing the development of expert systems rarely used the term 'expert' to refer to human expertise. Although, by definition, expert systems reference some human capacity or ability that a system is intended to emulate, the frequent lack of discussion of human expertise or involvement of an identified expert made it impossible to assess any basis for defining expertise.

In our search, we focused on papers that involved the development of a ML system, including papers that only focused on components of ML systems, such as ontologies or causal maps. The publications removed from our search results included entire books, PhD theses, magazine articles, review papers, and position papers. After removing papers that did not meet our inclusion criteria we were left with 112 papers. For a complete listing of the papers in our corpus, reference Table \ref{publishers} in the Appendix. Next, we conducted a second full coding pass of the 112 papers. We again randomly split publications among the two researchers and designated 10\% of the papers to be coded by both authors. We met periodically to discuss ambiguities, refine, and collapse codes until both researchers found agreement on a final set.

\subsection{Methodological Limitations}
Importantly, ``expert'' is not a discrete category-- rather it exists on a spectrum. Thus, our binary keywords (i.e., ``expert'' and ``non-expert'') may not capture instances where authors view stakeholders as highly knowledgeable unless they are described as experts explicitly. We focused on references to expertise in paper titles and paper metadata. Thus, our search also excluded a range of publications in which a relevant search term did not appear in the title of a paper but where relevant mentions or discussions of expertise appeared. However, we scoped our search to focus on explicit invocations of ``expert'' terminology to highlight publications which treat questions of expertise as thematically central to the research so as to gather a more concrete signal. Even using our chosen methodology, a number of publications only mentioned expertise in their title. We expect that many definitions of (non-)expertise are implicit across a range of works, thus we treat our findings as a prototypical snapshot rather than a comprehensive account of how ML researchers discuss expertise. In addition, focusing on paper titles helped to keep the volume of search results tractable and enabled us to leverage the breadth of publications that dblp.org afforded. We also felt that focusing on papers that mentioned our search terms in their titles was in keeping with our goal of investigating ML research in which there was explicit discussion and reference to expertise central to the narrative of the work. In addition, our search was limited to publications in English and more likely to reflect Western research practices, though we note that a number of publications were translated from other languages.
\section{Findings}
Through our iterative analysis, we developed codes to characterize 1) the basis on which expertise is determined, 2) the phase of development in which experts or non-experts were engaged, and 3) the role experts and non-experts played in development. Next, we present these codes as a taxonomy and discuss how often they occur.

\subsection{Overview}
In general, there was a tendency for publications to only vaguely describe who or what constituted expertise for the development of a given system, when one was given at all. Nearly half, or 51 publications, did not provide explicit criteria that characterized the experts or non-experts they engaged. Non-expertise was more often explicitly defined and tended to be engaged in more limited development roles. Experts and non-experts were typically engaged to provide data in the form of annotations or domain-specific facts or definitions for ontology development, to validate labels or model predictions, or to participate in end-user evaluations. 98 of 112 publications made reference to experts and 30 made reference to non-experts. Code counts total to more than 112, as some publications described the involvement of both experts and non-experts in various capacities or involved the same expert or non-expert across multiple steps of development. 16 publications discussed both experts and non-experts.

In the publications we surveyed, experts and non-experts were engaged at similar rates in each development phase and in their roles with the exception of the phase of model evaluation and the role of end-user. Non-experts were more often engaged in model and system evaluation and as system end-users. This is in line with the range of computer science research focused on supporting novice users.

\begin{table*}[ht]
\begin{center}
\renewcommand{\arraystretch}{1.2}
\begin{tabular}{ p{.19\linewidth} | p{.065\linewidth} | p{.66\linewidth}}
 \textbf{Expertise Basis} & \textbf{Expert Count} & \textbf{Description} \\ \hline
 
 \textbf{None Given} & 51 (9) & No explicit definition of expertise provided that is not circular (e.g., ‘an expert in X field’) and that is more specific than the general notion that an expert has relatively more knowledge than others in a given domain. \\ \hline
 
 \textbf{Education/\linebreak[4]Certification} & 30 (20) & Defined or implicitly determined based on educational attainment or certification  \\  \hline
 
 
 \textbf{Profession} & 31 (7) & Defined or implicitly determined based on a job held \\ \hline

 \textbf{Work experience} & 20 (11) & Defined on the basis of time spent working in a field, setting (e.g., intuition developed over time rather than rote textbook memorization of concepts), OR with a tool (e.g., animation software) \\ \hline
 
 
 \textbf{Human ability} & 1 (1) & Defined or implicitly determined on the basis of common human ability (e.g., ability to pick up objects) that machines lack or for which they have limited proficiency \\ \hline
  
 \textbf{Official source} & 3 (0) & Explicitly defined or implicitly determined on the basis of being a part of official references, documentation, or law (typically used in cases where knowledge is obtained via automated extraction) \\ \hline
\end{tabular}
\caption{Codes describing the bases for identifying "experts" referenced in ML publications. "Non-expert" counts in parentheses. Non-expert code counts were based on the lack of such qualifications.}
\label{tab:basis}
\end{center}
\end{table*}

\begin{table*}[t]
\begin{center}
\renewcommand{\arraystretch}{1.2}
\begin{tabular}{ p{.19\linewidth} | p{.065\linewidth} | p{.66\linewidth}}
 \textbf{Development Phase} & \textbf{Expert Count} & \textbf{Description} \\ \hline
 
 \textbf{Data Collection} & 44 (5) & Includes data collection process decisions (e.g., selection of label schema) as well as participation in primary data collection (e.g., annotation; providing source data) \\ \hline
 
 \textbf{Algorithm Design} & 15 (1) & Decision steps regarding algorithm features (e.g., determining Bayesian probabilities; selecting predictive variables), including:  \\
 
 \hspace{5mm} \textit{Variable Selection} & 4 (0) & Process of determining variables/features relevant for modeling \\
 
 \hspace{5mm} \textit{Knowledge Rep.} & 20 (4) &  Creation of a knowledge representation (e.g., knowledge base, causal map, ontology) \\ \hline

 \textbf{Evaluation} & 26 (14) & Evaluation apart from accuracy testing, including in-context assessments such as user and user experience evaluations \\ \hline
 
 \textbf{Testing} & 15 (4) & Testing or validation of a system through quantitative metrics, such as accuracy, F1, or ROC-AUC measurement (e.g., providing ground truth annotations for accuracy measurement; verifying the accuracy of system outputs in evaluative settings) \\ \hline
  
 \textbf{Problem \linebreak[4] Formulation} & 4 (0) & Determinations about what an algorithm should predict and how it should be used \\ \hline
 
 \textbf{None} & 17 (11) & (Non-)Expert played no direct role in the development process \\ \hline 
\end{tabular}
\caption{Codes describing the phases in which Experts and Non-Experts were engaged. "Non-expert" counts in parentheses.}
\label{tab:phase}
\end{center}
\end{table*}

\subsection{Basis of Expertise}
Table \ref{tab:basis} shows the full set of codes describing the bases for definitions of expertise, however we note some details here. “None Given” refers to publications in which no explicit definition of expertise is provided, of which there were 51. In particular, we considered this code applicable when a publication references engagement with an expert or non-expert but does not provide any specification beyond the reasonable assumption that an expert wields some kind of specialized knowledge in a domain. When we coded a paper as “None Given,” we also added an additional code describing a basis for expertise that we could infer from research design. For example, \cite{clark2003enabling} did not provide an explicit definition for expertise, but referred to participants in their user evaluation as ``experts in biology''. The participants in the evaluation were 3 graduate students in biology as well as one undergraduate student in biology. Therefore, we concluded that college-level coursework in the domain was a primary criterion for determining expertise. Although the nature of an individual’s expertise could often be inferred to some degree, we found that in some cases, the lack of specificity introduced challenges for scientific reproducibility. However, it is worth noting that for 29 papers, there was no obvious definition that could be derived. Implied definitions most often were inferred from information about the examples, participants, or annotators mentioned as "expert".

The next most common code for the basis of expertise was “Education/Certification”. This included publications that defined expertise on the basis of an advanced degree or training to use a system or tool. Unsurprisingly, this code frequently appeared alongside “Profession” and it is worth noting that, while some publications specifically emphasized work experience rather than certification (e.g., \cite{nishino2017fault} specifically engaged expert engineers with more than 10 years of machine maintenance experience), profession typically entailed a certification or education in the domain of interest (e.g., \cite{vaidyanathan2011using} specified Board-certified dermatologists as experts. Much less common were "Human Ability" and "Official Source". "Human Ability" described definitions based in common human capacities, such as the ability to pick up and hold objects for recordings used by \cite{shahri1997neuro} in training robot arms, and the ability to identify and distinguish voices, which \cite{greenberg2011including} relied on for evaluating a speaker recognition system. "Official Source" describes expert input, data, or information that came from formal documentation, such as policy or guidance, created by an unspecified group of experts \cite{steenwinckel2021flags}.

\subsection{Development Phase}
Table \ref{tab:phase} shows the codes describing the phases of development in which experts and non-experts were engaged. With regard to when experts and non-experts were engaged, 39 distinct publications described efforts in which experts or non-experts were engaged in data collection, which typically involved data annotation or the provision of source data, such as in \cite{oosterlinck2020one}, which asked experts in fraud detection to generate and validate synthetic data representing fraudulent subscriptions to a telecommunications family plan. 43 unique publications described experts involved in algorithm design and related activities, which broadly included variable selection and contributing to knowledge representations, such as ontologies or causal maps. Experts often contributed textbook knowledge or validated ontological information, such as in \cite{gordon2015combining} where experts evaluated an ontology for adherence to medical guidelines.

Experts were less often engaged in feature engineering or in determining relevant information for prediction. Another 33 publications involved experts and non-experts in end-user evaluations, such as user studies, or validation of model predictions through ground-truth testing data, respectively. For example, \cite{gholami2018replicating} engaged experts to analyze waveform patterns representing patient breathing on ventilators and to provide ground-truth to test a predictive algorithm designed to detect mismatches between mechanical ventilator delivery and patient respiratory needs.

\subsection{Expert Role}
Table \ref{tab:role} shows the codes describing the various roles experts and non-experts played in development. Experts and non-experts were engaged as source data providers 52 and 11 times, respectively, across 45 unique publications. This code refers primarily to data annotation and a few instances where experts and non-experts were the subject of data collection, such as in \cite{vaidyanathan2011using} where medical experts' eye movements and fixations were recorded and used as evaluation data to determine whether image processing algorithms were able to identify perceptually-relevant image regions. In another 24 unique publications, experts were engaged as end-users 15 times and non-experts were engaged as end-users 16 times in a system evaluation.

Interestingly, in 13 instances of discussion of experts and non-experts, across 12 papers, experts and non-experts were not engaged in development or evaluation at all. 7 of these publications described the development of systems designed to support experts or non-experts in analyses or decision-making. For these papers the expert or non-expert was conceptualized as an end-user; but, because there was no direct or indirect involvement in development, we did not label these as "End-User". In another 3 instances, the role of experts or non-experts was completely ambiguous, but no direct engagement was described. The remaining 3 publications involved processes of automated information extraction from documents or other sources deemed to be "expert" or "non-expert", such as a dataset of Vietnamese text scraped from news sites and online forums that was described as "non-expert" because it lacked phonetic and phonological linguistic information typically used in automated speech recognition for higher-resource languages \cite{luong2016non}. These are typically cases where it is deemed more efficient or easier to automate knowledge extraction from documents rather than manually consult with an expert.
\section{Discussion}

\begin{table*}[ht!]
\begin{center}
\begin{tabular}{ p{.19\linewidth} | p{.065\linewidth} | p{.66\linewidth}}
 \textbf{Expert Role} & \textbf{Expert Count} & \textbf{Description} \\ \hline
 
 \textbf{Source Data Provider} & 37 (10) & Provides data used directly in model training or testing (e.g., ground truth, label set, training examples), for example:  \\
 
 \hspace{5mm} \textit{Data Subject} & 8 (1) & Is observed or recorded (e.g., sensors record Non/Expert movements such as eye-fixation points) \\
 
 \hspace{5mm} \textit{Expert Data} & 7 (0) &  Creates official source that is scraped (e.g., official website) \\ \hline
 
 \textbf{End User} & 29 (16) & Interacts with system as a human-in-the-loop participant or supports end-user evaluation, typically through a usability assessment \\ \hline

 \textbf{Informant} & 32 (3) & Is used as a reference for explaining or defining features of an established process,typically in a one-way manner (e.g., defines the properties of a biological process). This includes determining the validity of some aspect of system development (e.g., verifying that synthetically-produced data is realistic) \\ \hline
 
 \textbf{Consultant} & 18 (0) & Provides information or insight about the problem or context at hand, but is not directly involved in design or creation of model components, such as knowledge graph creation or variable selection. (e.g., providing recommendations for reliable data sources; acting as interviewee to explain how an existing process works) \\ \hline
 
 \textbf{No Direct Role} & 7 (7) & Source produced by an expert is consulted, but expert is not, similar to ``expert data'' (e.g., textbooks policy defined by experts). \\ \hline
 
 \textbf{Ambiguous} & 4 (2) & Not enough information is provided to determine how a Non/Expert contributed to system development \\ \hline

 \textbf{Designer} & 4 (0) & Actively involved in decisions about what the ML system should do or how the algorithm should operate (e.g., variable selection, active decisions about what is included in an ontology–not simply providing definitions or providing textbook knowledge) \\ \hline
 
\end{tabular}
\caption{The codes describing the roles in which Experts and Non-Experts were engaged. "Non-expert" counts in parentheses.}
\label{tab:role}
\end{center}
\end{table*}

\subsection{Datafication and De-skilling}
In our review, the the roles experts played tended to rely on an implicit assumption that expertise can be easily extracted and divorced from experts, echoing prior work characterizing ML development \cite{forsythe1993engineering}. We see 84 instances total in which experts were asked to provide source data for training or testing, or in which experts were engaged as informants. Across these engagements, experts provided inputs to development efforts with limited insight and influence over system goals. This implicitly treats expert knowledge as modular and decontextualized.

\subsubsection{Datafication} Processes of datafication, or ``the transformation of qualitative behavior and tacit knowledge into quantified actions and codified data’’ \cite{fischer2021datafication} – shape which aspects of knowledge are incorporated into ML systems. However, our findings point to an explicit interest on the part of ML researchers in engaging aspects of expertise that may defy quantified knowledge capture through engagements with individual experts outside of domain contexts. There were 26 instances in which expertise was determined on the basis of work experience in addition to or instead of educational background or certification. For example, \citet{schamoni2019leveraging} focused on experts with clinical experience managing patients with sepsis, explicitly asking experts to make decisions in simulations based on experience rather than pre-defined medical guidelines. These publications point to an acknowledgement that protocols and standard procedures are not always sufficient for making expert decisions. This is implicitly acknowledges the importance of experiences outside of formal educational training and testing, which are varied and difficult to quantify. As \citet{mejias2019datafication} argue, ``something that cannot be codified as a potential network member cannot be accounted for''. Ways of knowing that cannot be quantified do not shape the information produced by algorithmic systems. This includes the collective knowledge and interactions among actors and the infrastructure surrounding experts. The social embeddedness of expertise contributes to the ability of human decision-makers to reflect on decision boundaries before making decisions for marginal or novel cases, whereas algorithmic systems can only incorporate feedback after producing a decision or receiving input from human review \cite{alkhatib2019street}.

In practice, an expert acts as more than ``repository of knowledge'' \cite{handy1993}; they interpret and apply knowledge in contexts comprised of other experts, stakeholders, organizational structures, technologies, and information flows. In addition, other forms of knowledge, such as tacit knowledge, shape expert decision-making and how knowledge is applied. Complexities of the social contexts in which experts operate are difficult to capture and model, including elements of expertise that the expert, ``does not know he knows'' \cite{collins1986s}. These elements are perhaps most obviously missed when experts are asked to provide ground truth annotations without engagement in other aspects of system design, which we frequently observed. Overall, experts were engaged in more than one phase of development in just 18 papers. Although quantification is fundamental to ML development, expert input on goals and design provides an opportunity to check assumptions and shape system application. Our findings point to 22 instances where experts were engaged in this way, and just 4 instances where they were engaged in problem formulation. We see potential to integrate community-based participatory research (CBPR), which not only challenges assumptions about what constitutes expertise, but also allows researchers and experts to co-create systems and goals that meet the needs of stakeholders.

\subsubsection{Deskilling} Our findings point to patterns of what other scholars have observed regarding deskilling-- or engaging experts in ways that ignore or devalue their breadth and depth of knowledge \cite{sambasivan2022deskilling, zuboff1988age}. In ML development, deskilling has affected a variety of experts, ranging from farmers \cite{sambasivan2021re} to medical doctors \cite{duran2021deskilling}. Critiques of automation more broadly point to ways in which it refocuses workers on menial tasks \cite{zuboff1988age}, rendering traditionally valued aspects of expertise, such as tacit knowledge, unused. We see this reflected in how often experts were engaged in activities outside of design or problem formulation. \citet{sambasivan2021re} point out a tension between ML goals to emulate and extend human expertise and ML practices that devalue that same expertise in humans. Further complicating this tension are the social roles ascribed to the people that engage in these tasks. For example, they observe limited direct engagement between AI developers and a reductionist view of field workers as data collectors.

We also observe limited engagement in the type of knowledge experts are asked to provide and share. This underscores tensions in participatory ML development articulated by \citet{birhane2022power}, wherein participation toward improvements in algorithmic performance is typically undertaken within conditions set by the researchers and in one-off engagements. These engagements can limit experts' agency and influence on project directions and impacts, thereby devaluing other expertise they could bring to bear on a project. As stakeholders often outside of a project's origination, experts can already exert confined influence that stands to become more limited through deskilling. While most of the papers in our review deal with experts in professions with relatively good pay and social status, it is worth noting that pay or other forms of compensation for participation were rarely mentioned in the publications we reviewed. Although experts who occupy positions with relative social power are less vulnerable to exploitative dynamics compared with other stakeholders, deskilling is reflected in the extent to which expertise is not specified at all. 

A related question to the issue of how knowledge is valued and captured, is the question of what knowledge \textit{does} in relation to ML infrastructure. Each role that experts were engaged in exists along a spectrum of the degree to which it exerts influence over the ML infrastructure it produces. For example, Source Data Providers create the data upon which development teams apply infrastructure paradigms or that teams may mold into established infrastructure, such as through benchmark datasets \cite{denton2020bringing}. On the other hand, Algorithm Designers play a more integral role in shaping how data should be interpreted and used. This is not to suggest that Designers are in a position to completely subvert infrastructural norms in the field; however they are better able to direct how their knowledge can or should be utilized. Thus, we must consider engagements with experts in relation to the degree to which they are able to exert influence over the artifacts they are asked to produce.

\subsection{Recognizing Expertise in Support of Responsible AI Practice}

The issues of \textit{who} is recognized as expert and how to incorporate a wider range of knowledge into AI fall within the domain of RAI practice. RAI must develop practices to identify expertise in robust ways and document decisions about the expertise engaged. This need is echoed in recent work focused on characterizing the identities and perspectives of data annotators in relation to dataset development processes and goals \cite{diaz2022crowdworksheets}.

\subsubsection{Which Knowledge is Recognized} When the basis for expertise was explicitly named among the publications we reviewed, there was a tendency to name and recognize expertise in terms of education and professional standing, which we saw in 42 publications in total. For many systems and applications, a need for specialized knowledge obtained through specific training or education is intuitive. At the same time, standpoint theory describes acquired knowledge, both formal and informal, as rooted in “a situated, embodied, location in the world” \cite{harding1986science}. Thus, knowledge is shaped experientially and can produce unique insights that formal learning environments may not enable. In this frame, knowledge is informed by embodied experience, rather than as an objective truth or "view from nowhere".



For example, \citet{patton2019annotating} compared social work graduate students to community members in an annotation task aimed at identifying Twitter content that could predict offline violent interactions. The community member annotators were able to identify details related to gang interactions that the graduate students did not recognize despite formal training in social work and task training from community members. While it is unclear how these differences in annotation may have influenced model performance, it is a clear and relatively rare example of documented differences in ML development that arise from expertise in the same domain derived from different knowledge sources. 

\subsubsection{Reflecting on Expertise} Ultimately, there is a need to reflect on the expertise desired by system developers and the expertise engaged and represented in development. Beyond acknowledging forms of knowledge gained in experiential and nontraditional ways, we argue that the labels ‘expert’ and ‘expertise’ must be unpacked as a matter of power, because they direct attention to specific sources and figures as knowledgeable and canonical. As Handy posits, "[e]xpertise provides expert power vested due to acknowledged expertise” \citet{handy1976so}. In other words, recognition of expertise confers power and influence upon those acknowledged.

Identifying a limited, canonical set of experts and extrapolating from the input they provide risks ignoring other perspectives or forms of knowledge, particularly when the expertise used is presented as universally applicable. This is reflective of epistemic trespassing, which occurs when ``thinkers who have competence or expertise to make good judgments in one field, [...] move to another field where they lack competence—and pass judgment nevertheless'' (\cite{ballantyne2019epistemic}, p. 2)." In this context, epistemic trespassing functions as a form of epistemological discrimination, wherein stakeholders possessing nontraditional and minoritized forms of knowledge in a given domain are sidelined in favor of individuals whose expertise primarily resides in another domain. As \citet{handy1993} describes, there is ``...a complex system of concepts and beliefs concerning the nature of expertise, the appropriate course of action to follow when one experiences a lack of expertise, the appropriate person to obtain it from, and so on.'' We extend this to include concepts and beliefs that shape disciplinary norms, new and old, such as those underpinning calls in participatory AI to acknowledge the expertise of marginalized communities. This line of critique is not to paint standardized benchmarks derived from traditional experts as wholly bad or unusable. Instead, we are concerned that both the logics that inform expert selection and engagement and the values that chosen experts apply are fated to become buried in the data and technologies that serve as current and future infrastructure, regardless of methodology. Even work in participatory AI relying on HCI experts to liaise with and speak on behalf of impacted stakeholders, risks effectively using formal expertise to shut out marginalized expertise \cite{delgado2023participatory}.

Failing to explicitly attend to the relevant aspects of identified expertise risks perpetuating representational harms \cite{crawford2021atlas}. This results from failing to acknowledge that within a single domain there can exist variations in expert practice and opinion rooted in cultural differences as well as by presenting an underspecified, normative perspective as canonical. Many researchers have named reasons for legitimate variation in ground truth judgments, for example, differences in social experience and cultural context \cite{davani2021dealing, diaz2018addressing, basile2021toward}. A simple yet critical step for recognizing the value of domain expertise lies in naming the bases of identified expertise and how it is intended to contribute to development. This provides transparency about which knowledge is setting the standard for performance, enables downstream users to determine whether that knowledge is relevant to their task, as well as supports scientific reproducibility by specifying which expert inputs informed development so that follow up work can validate findings.

\subsection{Implications for ML Research}
We conclude by urging ML developers to recognize expertise more expansively and afford experts expanded opportunities to shape the infrastructural artifacts that their engagements produce. Based on this work, we implore researchers developing ML systems in collaboration with domain experts to make the following conceptual considerations. Drawing from \citet{dourish2006implications}, we stress the importance of finding conceptual clarity in collaborative goals and motivations before engaging with domain experts. While we focus on ML, these considerations can also be applied to other fields to combat epistemological discrimination.

\subsubsection{Recognizing Expertise} Recognizing expertise more broadly must entail honoring marginalized expertise, which often takes the form of knowledge gained through means outside of formal or gate-kept institutions. At a high-level, we must reflect on the goals of invoking the language of "expertise" when describing and structuring ML development. Input from those identified as experts is typically prioritized because "expert" communicates that a person has the most relevant knowledge about that domain. However, concerns about fairness and equity behoove input from stakeholders who may not know the intricacies of technological architecture, but who possess deep knowledge of their own experiences with the impacts of technological systems as well as the social contexts with which researchers seek to align their systems. As a result, RAI development can consider different sets of stakeholders as expert in different stages of development. For example, linguists may provide expertise about language features, while community groups can evaluate cultural nuances in how language is understood in their communities.
\begin{itemize}
    \item What forms of expertise do identified domain experts have and how does this knowledge compare or contrast with others who are knowledgeable about this domain?
    \item Are there others who may wield relevant knowledge and that can be engaged with who may not be considered experts by traditional, certification-focused measures?
    \item How might differences in expertise across sources shape the dataset, model, or other artifact under development?
\end{itemize}

\subsubsection{Engaging Expertise} As we push for an expanded set of people to be included in data work, among other aspects of ML development, we look to scholarship on ethics in data labor to inform how experts can be more responsibly engaged. There are opportunities through what \citet{desmond2021increasing} term iterative and improvisational design of ground truth. These practices begin from an acknowledgement that an initial label set may not ultimately be usable, or ideal, such as in their given example of developing labels from physicians' clinical notes. Arriving at a suitable schema of labels required that the research team analyze source data in different ways, develop labels, and align among team members about their interpretations. Such a process enabled experts to shape the labeling paradigm and what it captured. At the same time, it required that the physicians they worked with were given a thorough understanding of system goals and components. Though effortful, involving experts more robustly in the design process serves to give experts insight into development goals and allowed them to play a role in defining how knowledge is captured, interpreted, and used.
\begin{itemize}
    \item How do different modes of engagement restrict or enable experts in the kind of knowledge they can share?
    \item In what ways might development rely on knowledge that is particularly difficult to elicit, such as tacit knowledge?
    \item Is there knowledge or insights that could be better elicited through non-individualized engagements (e.g., focus groups; collaborative activities with other experts)?
    \item How might the development plan involve domain experts in ways that engage their skills beyond querying them as a human textbook or repository of knowledge?
\end{itemize}

\subsubsection{Documenting Expertise} Documenting disparate motivations for inclusion can serve to mitigate issues of epistemic trespassing. Domain experts are not epistemic trespassers in that they have no legitimacy in sharing perspectives. Rather, it is that certain kinds of knowledge in a subject area are repeatedly engaged while others are not. In this way, identified experts' knowledge may be over-relied upon to make decisions when other perspectives may also complement development goals. In this vein, \citet{smart2020why} use reliabilism to critique erroneous trust in algorithmically-produced knowledge. The theory states that ``a belief is warranted if it has been produced by a reliable process or method.'' However, \citet{smart2020why} argue that the black box nature of ML models means that perceptions of reliability can often be misplaced. Thus the knowledge sources embedded in systems can be poorly understood or buried.

Transparency artifacts bring reflection to how and whether different types of expertise support development goals. In addition, clarity around expert sources improves scientific reproducibility. As in human subjects research, reporting sampling criteria enables peer scrutiny of scholarship and potential biases that study design can introduce.
\begin{itemize}
    \item What explicit documentation decisions being made about the knowledge and expertise being engaged with?
    \item How does resulting documentation enable or restrict peer review of research design and data integrity?
\end{itemize}

Ultimately, RAI must be concerned with operationalizing responsible and sustainable practices of engaging expertise, including who is engaged as an expert, how they are engaged, and how their input is utilized. Based on a systematic review of ML publications, we contribute a taxonomy detailing how expertise is characterized, the roles that experts and non-experts play in development, and when in development they are typically engaged. On the backdrop of calls for expanded participation in ML and AI development, we discuss trends in ML research in relation to how knowledge held by marginalized groups is devalued and ignored and the need for more equitable recognition of expertise toward responsible ML practices. Finally, we provide provocations to aide RAI researchers and developers in clarifying goals and motivations before engaging with domain experts.

\bibliography{aaai24}

\section*{Appendix}
\begin{table*}[ht]
\centering
\begin{tabular}{@{}c|cccccccc@{}}
\multicolumn{1}{l|}{\textbf{}} &
  \textbf{\begin{tabular}[c]{@{}c@{}}Data\\ Collection\end{tabular}} &
  \textbf{Testing} &
  \textbf{Eval.} &
  \textbf{\begin{tabular}[c]{@{}c@{}}Alg.\\ Design\end{tabular}} &
  \textbf{\begin{tabular}[c]{@{}c@{}}Var.\\ Selection\end{tabular}} &
  \textbf{\begin{tabular}[c]{@{}c@{}}Knowledge\\ Rep.\end{tabular}} &
  \textbf{\begin{tabular}[c]{@{}c@{}}Problem\\ Form.\end{tabular}} & \textbf{None} \\ 
{\textbf{Designer}}                                & 1                          & 0 & 1                          & 1 & 0 & 0 & 1 & 0 \\
\textbf{Consultant}                                                     & 3                          & 2 & 3                          & 5 & 0 & 3 & 2 & 0 \\
\textbf{Informant}                                                      & 9  & 6 & 3                          & 6 & 1 & 7 & 0 & 0 \\
\textbf{\begin{tabular}[c]{@{}c@{}}Source Data\\ Provider\end{tabular}} & 23 & 6 & 1                          & 1 & 2 & 4 & 0 & 0 \\
\textbf{Data Subject}                                                   & 5                          & 0 & 0                          & 0 & 1 & 0 & 0 & 2 \\
\textbf{Expert Data}                                                    & 2                          & 0 & 1                          & 1 & 0 & 0 & 0 & 3 \\
\textbf{End-User}                                                       & 1                          & 1 & 18 & 1 & 0 & 2 & 1 & 5 \\
\textbf{Ambiguous}                                                      & 0                          & 0 & 0                          & 0 & 0 & 2 & 0 & 2 \\
\textbf{None}                                                           & 0                          & 0 & 0                          & 0 & 0 & 2 & 0 & 5 \\ 
\end{tabular}
\caption{The co-occurrences of expert roles (on the left) and development phases (on the top) showing how often an expert in a given role was engaged in a corresponding development phase. The most frequent co-occurrences are highlighted.}
\label{tab:co-occurrences}
\end{table*}

\end{document}